\documentclass{article}
\usepackage[table]{xcolor}
\usepackage{graphicx}
\usepackage{epstopdf}
\usepackage{bm}
\usepackage{adjustbox}
\usepackage{array}
\usepackage{float}

\usepackage{booktabs}                                   
\usepackage{multirow}                                   
\usepackage{makecell}                                   
\usepackage{tablefootnote}                              
\usepackage[symbol]{footmisc}                           
\renewcommand{\thefootnote}{\arabic{footnote}}          
\usepackage{amsmath,amssymb}                            
\usepackage{xcolor}                                     
\usepackage{enumitem}                                   
\usepackage{subcaption}                                 
\usepackage{stfloats}                                   

\usepackage{wrapfig} 

\usepackage[bookmarks=true]{hyperref}     
\usepackage{xspace}                       

\setlist[itemize]{leftmargin=20pt} 

\newcommand{\eat}[1]{}                                  

\newcommand{\ours}{SwitchVLA\xspace}

\usepackage{multirow}
\usepackage{booktabs}

\usepackage{algpseudocode}  
\usepackage{algorithm}
\usepackage{amsmath}

\usepackage{mathtools}
\usepackage{hyperref}
\usepackage{enumitem}

\usepackage{hyperref} 

\usepackage[preprint]{corl_2025} 

\title{\ours: Execution-Aware Task Switching\\for Vision-Language-Action Models}

\definecolor{linkcolor}{rgb}{1,0.5,0} 

\author{
 \normalfont
 Meng Li$^{1,*}$, Zhen Zhao$^{1,*}$, Zhengping Che$^{1,*}$, Fei Liao$^{1}$,\\
 Kun Wu$^{1}$, Zhiyuan Xu$^{1}$, Pei Ren$^{1}$, Zhao Jin$^{1}$, Ning Liu$^{1}$, Jian Tang$^{1,\dagger}$\\
 $^1$Beijing Innovation Center of Humanoid Robotics\\
 \href{https://switchvla.github.io}{\textcolor{linkcolor}{https://switchvla.github.io}}
}

\begin{document}
\maketitle

\renewcommand{\thefootnote}{ } 
\footnotetext{
$^{*}$Equal contribution.
$\dagger$Corresponding author.\\
\phantom{XXX}{\texttt{\{gary.li, alex.zhao, z.che, leofly.liao, gongda.wu, eric.xu, chris.ren, mustafa.jin, neil.liu, jian.tang\}@x-humanoid.com}}
}

\begin{abstract}
Robots deployed in dynamic environments must be able to not only follow diverse language instructions but flexibly adapt when user intent changes mid-execution.
While recent Vision-Language-Action (VLA) models have advanced multi-task learning and instruction following, they typically assume static task intent, failing to respond when new instructions arrive during ongoing execution. 
This limitation hinders natural and robust interaction in dynamic settings, such as retail or household environments, where real-time intent changes are common.
We propose \textbf{\ours}, a unified, execution-aware framework that enables smooth and reactive task switching without external planners or additional switch-specific data. 
We model task switching as a behavior modulation problem conditioned on execution state and instruction context. 
Expert demonstrations are segmented into temporally grounded contact phases, allowing the policy to infer task progress and adjust its behavior accordingly. 
A multi-behavior conditional policy is then trained to generate flexible action chunks under varying behavior modes through conditioned trajectory modeling. 
Experiments in both simulation and real-world robotic manipulation demonstrate that \ours enables robust instruction adherence, fluid task switching, and strong generalization—outperforming prior VLA baselines in both task success rate and interaction naturalness.
\end{abstract}

\keywords{Imitation Learning, Vision-Language-Action, Task Switching}

\section{Introduction}
\label{sec:introduction}
Vision-Language-Action (VLA) models have gained increasing attention for their ability to map natural language instructions and visual observations into executable robotic actions. Recent progress has significantly advanced instruction following and multi-task learning capabilities~\citep{brohan2022rt,bharadhwaj2023roboagent,chi2023diffusion,goyal2023rvt,zhang2024hirt,fu2024mobile,cheang2024gr,kim2024openvla,black2024pi_0,wu2025momanipvla,kim2025fine}. However, most of these models operate under the assumption that task intent remains fixed throughout execution. As a result, they struggle to respond when new instructions are issued mid-task—an assumption that falls short in real-world, interactive environments. For example, service robots in retail settings must flexibly adapt to customer hesitation, last-minute exchanges, or item returns that arise during ongoing interactions.

Despite its practical importance, the ability to dynamically respond to changing task intent has mainly been overlooked in prior work. Supporting such behavior requires VLA models to go beyond static instruction-to-action mapping. They must continuously track execution progress and reason about when to forward, rollback, or advance ongoing behaviors in response to changing user input. Crucially, this must be achieved while maintaining multi-task generalization, enabling robots to fluidly adapt across task boundaries without task-specific heuristics or handcrafted transitions.

\begin{figure}[t]
\centering
\includegraphics[width=0.99\textwidth,height=0.25\textwidth]{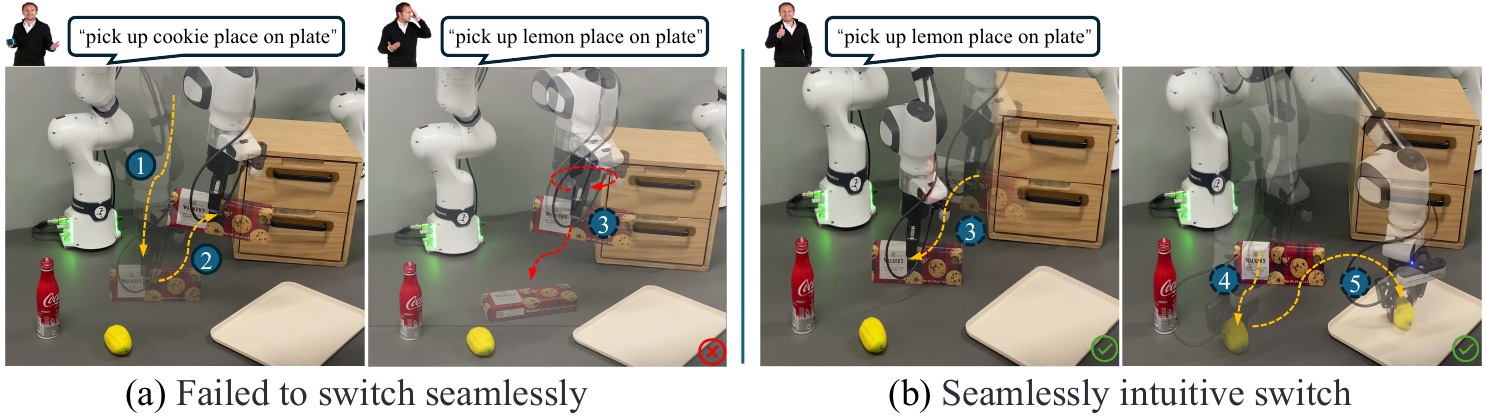}
\caption{
  \textbf{(a)} Processes 1 and 2 show normal task execution. 
  When the user changes their mind (e.g., ``pick up lemon place on plate''), conventional VLA models cannot adjust its plan, leading to erratic behavior like oscillation or dropping items, as seen in Process 3.
  \textbf{(b)} A more natural response involves returning the previously held item (e.g., placing down the cookie box in Process 3 and then picking up the lemon and placing it on the plate in Processes 4 and 5).
}
\vspace{-5pt}
\label{fig1}
\end{figure} 

Some recent approaches leverage large language or vision-language models for high-level task planning and interactive intent grounding~\cite{yuan2023plan4mc, namasivayam2023learning, huang2024rekep, liu2024okrobot, shi2025hi}, enabling re-planning when a new instruction is issued. However, these methods typically lack low-level execution flexibility: even if a task switching is inferred, the robot must wait for the current task to complete before re-planning, limiting real-time adaptability. Moreover, inference latency in large models makes high-frequency reactive reasoning infeasible. Other works attempt to enhance failure recovery and task switching by collecting additional demonstrations~\citep{dai2024racer,huang2025adversarial}, yet this strategy scales poorly—switching behaviors grow combinatorially with task diversity, making exhaustive data collection impractical. Therefore, a key challenge remains: \textit{how to endow VLA models with the ability to smoothly and reactively switch tasks during execution using only existing trajectory data, without high-level planning or extra demonstrations.}
As shown in Fig.~\ref{fig1}, conventional VLA policies often struggle to naturally handle dynamic human instructions, such as discarding or mishandling objects when the target changes mid-task.
In contrast, the expected model behavior enables intuitive responses by automatically executing recovery actions before proceeding with new commands—a behavior pattern closely mirroring natural human operation.

To address the limitations of prior work, we propose \textbf{\ours}, a unified execution-aware framework that enables dynamic behavior switching within a VLA policy. 
Instead of relying on external planners or manual scripts, \ours gives the model fine-grained execution awareness, enabling it to reason about task progress and adaptively forward, rollback, or advance actions in response to changing instructions.
We formulate task switching as a state-conditioned behavioral modulation problem, where local execution dynamics drive continuous decision-making. 
To support this, we segment expert trajectories into time-based contact phases, enabling the policy to infer the current stage of the task and adjust its behavior accordingly.
By employing execution-aware conditional policies, SwitchVLA can select behavior modes and predict future actions without requiring additional switch-specific data. 
Empirical results show that \ours enables coherent transitions, improves robustness under dynamic task changes, and generalizes effectively across multi-stage manipulation tasks.

In summary, our contributions are as follows:
\begin{itemize}
  \item We propose \ours, a unified execution-aware framework that employs a novel training paradigm and innovative architecture to support dynamic task switching without relying on additional switch-specific data.
  \item We design a multi-behavior conditional policy capable of smoothly forwarding, rolling back, or advancing actions within a single policy backbone. 
  \item We empirically validate \ours in both simulated and real-world robotic manipulation tasks, demonstrating substantial improvements in task transition smoothness, recovery effectiveness, and instruction adherence over existing VLA baselines.
\end{itemize}

\section{Related Works}

\textbf{Vision-Language-Action Models for Robotic Control.}
VLA models have transformed robotic control by mapping multimodal inputs to actions~\cite{brohan2022rt,ahn2022can,chi2023diffusion,brohan2023rt,zhao2023learning,li2023vision,octo_2023,kim2024openvla,black2024pi_0,haldar2024baku,nasiriany2024rt,wu2025momanipvla,kim2025fine,contributors2025agibotworld,bjorck2025gr00t}. Many works integrate language with robotic actions via imitation learning, where language serves as either goal specification~\cite{brohan2022rt,brohan2023rt,kim2024openvla,black2024pi_0,kim2025fine,bjorck2025gr00t} or process-level planning~\cite{ahn2022can,hu2023look}.
While these methods achieve strong generalization and instruction-following via architectural and data-scale improvements, they often assume a static execution flow—treating instruction following as a single-shot action prediction task, without modeling execution dynamics or supporting real-time adaptation. 
For instance, OpenVLA~\cite{kim2024openvla,kim2025fine} and $\pi_0$~\cite{black2024pi_0} predict the next action from current observations and language but lack mechanisms for adjusting behaviors mid-execution. 
SayCan~\cite{ahn2022can} and ViLa~\cite{hu2023look} depend on external modules to replan when states diverge.

\textbf{Task Switching and Interactive Execution.} 
Recent work explores interactive execution with increasing focus on task switching, categorized into modular systems and learning-based methods.
Modular systems employ planners, hierarchical policies, or feedback loops~\cite{yuan2023plan4mc,namasivayam2023learning,hu2023look,huang2024rekep,liu2024okrobot,xiao2024robi}, but often rely on manually designed rollback, advance, or recovery strategies, limiting scalability and integration with end-to-end VLA models. They usually assume task completion before new instructions are issued.
For example, ReKep~\cite{huang2024rekep} focuses on high-level interaction but suffers from latency issues; YAY Robot~\cite{shi2024yell} supports context-aware correction but confines rollback to scripted demonstrations; Hi Robot~\cite{shi2025hi} enables multi-behavior learning but requires complex, behavior-specific training.
Learning-based methods~\cite{dai2024racer,belkhale2024rt,huang2025adversarial} support model-driven interaction but depend on additional specialized data. RACER~\cite{dai2024racer} introduces failure-aware recovery at the high level, relying on curated failure datasets, while ADC~\cite{huang2025adversarial} augments learning with extra demonstrations.

While most existing methods focus on high-level planning, they often neglect the challenges of dynamic execution and real-time task adaptation. In contrast, our work targets the low-level execution of VLA models, proposing a unified policy that enables smooth, reactive, and adaptive task switching during execution—without requiring additional demonstrations.

\section{Method}
\label{sec:method}
We introduce \ours, a unified learning framework for dynamic task-switchable action generation within Vision-Language-Action (VLA) settings. Our method treats task-switching not as a hard re-planning problem, but as a conditional behavior prediction challenge, where action dynamics adapt in response to evolving task intents and execution feedback.

\subsection{Problem Formulation}
\label{sec:problem_formulation}
\textbf{Standard VLA Execution.}
Given the robot's expert trajectory, $\tau = \{(l | o_t, q_t)\}_{t=0}^T$, where $l$ is the task language instruction for the trajectory, $o_t$ and $q_t$ represent the visual observation and robot state (e.g., joint angles) at time $t$, respectively. 
The goal is to learn a policy that maps $(l | o_t, q_t) \mapsto a_{t+1}$ in a behaviorally coherent manner, where $a_{t+1}$ denotes robot's action at time $t+1$. 

\textbf{Task Switching.}
In practical deployment, the robot may receive new task instructions $l'$ at arbitrary times during execution. Such dynamic inputs introduce out-of-distribution observation-instruction pairs $(l' | o_t, q_t)$, posing significant challenges for generalization. 
We identify two core sub-problems:
(i)~instruction grounding—aligning the policy with the latest instruction \(l'\), and
(ii)~execution-aware switching—using execution feedback (e.g., physical contact) to decide whether to forward, rollback, or switch to a new behavior mode.

To address these challenges in task switching, we introduce two auxiliary supervision signals—the contact state and the behavior mode—as key latent indicators of task phase and execution feedback.

\subsection{Supervisory Signals for Instruction-Aware Control}
\label{sec:supervisory_signals}

\textbf{Contact State.}
The contact state indicates physical interaction between the robot and objects. We define it as a binary variable $c_t \in \{0, 1\}$, where $0$ denotes no contact and $1$ denotes contact. It can be inferred through:
tactile sensing, gripper open/close signals, heuristic motion or force thresholds, or vision-language parsing using pre-trained models. This binary state evolves over time and informs task phase progression, guiding the system's next action strategy.

\textbf{Behavior Mode.}
\label{subsec:behavior_mode}
We define the behavior mode at timestep $t$ as $b_t \in \{0:\texttt{forward}, 1:\texttt{rollback}, 2:\texttt{advance}\}$, with each value corresponding to a distinct behavioral strategy: $\texttt{forward}~(b_t = 0)$ continues standard execution, $\texttt{rollback}~(b_t = 1)$ undoes previous actions upon detecting an intent mismatch while in contact, and $\texttt{advance}~(b_t = 2)$ transitions to a new subtask when the instruction updates and no physical interaction is present.

The labels of contact states and behavior modes can be weakly supervised from phase-aligned demonstrations or derived via behavior heuristics automatically parsed from execution feedback.
Together, the contact state $c_t$ and behavior mode $b_t$ provide key supervision signals to address the task-switching challenge.
The contact state offers real-time execution feedback to detect interaction phases, while the behavior mode encodes high-level task intent—whether to \texttt{forward}, \texttt{rollback}, or \texttt{advance}.
These signals condition the policy to adaptively align with updated instructions and respond coherently under dynamic execution.

\begin{figure}[t!]
\centering
\includegraphics[width=1.0\textwidth]{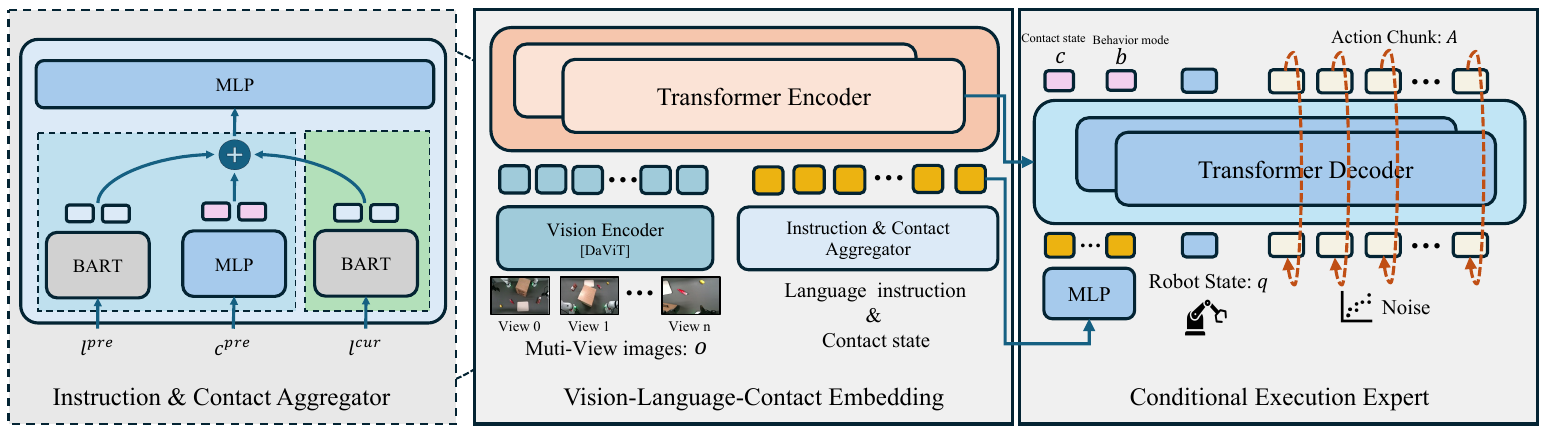}
\caption{
Overview of \ours. The framework consists of the Vision-Language-Contact Embedding module and the Conditional Execution Expert, which jointly fuse multimodal inputs to generate execution-aware and conditionally controlled actions.} 
\label{fig2}
\vspace{-5pt}
\end{figure} 
\subsection{Architecture Overview}
\label{sec:architecture_overview}
\ours establishes a unified architecture for robust and instruction-consistent task execution, as shown in Fig.~\ref{fig2}. The architecture consists of two core components: (i)~\textbf{Visual-Language-Contact (VLC) Embedding Module} encodes visual, language, and contact cues into unified representations. (ii)~\textbf{Conditional Execution Expert} decodes behavior-aware actions conditioned on the current multimodal embedding. We build \ours upon Florence 2~\cite{xiao2023florence}.

\subsubsection{VLC Embedding Module}
To represent execution context with temporal and semantic richness, the Visual-Language-Contact (VLC) Embedding module fuses multi-view RGB observations, contact-aware execution cues, and paired task instructions into a unified token sequence. 
This embedding serves as the core state representation for downstream behavior selection.

\textbf{Visual Encoder.}
We adopt a DaViT-based~\cite{ding2022davit} backbone to encode multi-view RGB observations $o$ into dense spatial tokens, projected for robust visual grounding.

\textbf{Instruction and Contact Aggregator.}
By integrating historical and current contextual information, a rich conditioning signal is formed to enable behavior-aware action generation. 
Specifically, the previous instruction $l^{\text{pre}}$ and contact state $c^{\text{pre}}$ reflect past intent and interaction history, and the current instruction $l^{\text{cur}}$ updates the semantic goal based on new task directives.
These tokens are embedded using BART for language and an MLP for contact encoding, then concatenated to form the conditioning input token sequence.

\begin{figure}[t!]
  \centering
  \includegraphics[width=1.0\textwidth]{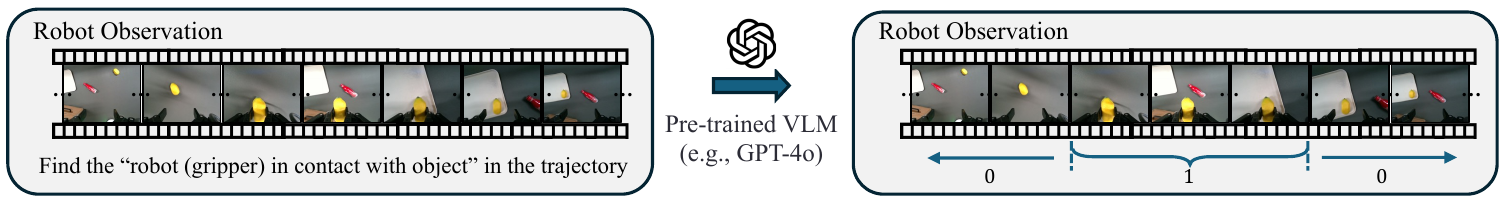}
  \caption{
    Identify and label time intervals of a specified event from trajectory data using a pre-trained VLM, such as GPT-4o. 
    For example, with the prompt ``Robot (gripper) in contact with object'', the model retrieves and labels the contact time intervals within the trajectory.
  } 
  \label{fig3}
  \vspace{-5pt}
\end{figure}

\textbf{Trajectory Parsing and Contact Annotation.}
In our implementation, we adopt a vision-language trajectory parsing approach: a pretrained VLM segments demonstrations into coarse, contact-aware phases based on wrist camera observations (Fig.~\ref{fig3}), with gripper open-close signals incorporated to enhance the reliability of phase boundaries. These lightweight annotations enrich the tokenized input without modifying the core learning pipeline.

\textbf{Transformer Fusion.} All tokens are fused using Transformer encoder-decoder~\cite{vaswani2017attention,xiao2023florence}, producing temporally and semantically rich embeddings.

\subsubsection{Conditional Execution Expert}
The Conditional Execution Expert module serves as a structured action decoder that integrates real-time contact signals with high-level behavioral intent to support adaptive and temporally coherent action generation. It simultaneously predicts three synchronized outputs in each timestep: the contact state $c_t \in \{0, 1\}$ reflecting whether the robot is in physical interaction with the environment, the behavior mode $b_t \in \{0, 1, 2\}$ indicating the current operational intent, \texttt{forward} execution, \texttt{rollback} of prior steps, or \texttt{advance} to a new subtask, and the action chunk $A_t = \{a_{t+k}\}_{k=1}^{K}$, which specifies a sequence of low-level actions spanning the next $K$ timesteps.

The structured prediction design facilitates contact-aware behavioral adaptation, enabling reliable mode switching based on interaction signals. It also supports explicit behavioral reasoning over \texttt{forward}, \texttt{rollback}, and \texttt{advance} actions, as well as smooth temporal transitions through joint decoding of multi-step action chunks—reducing jitter and enhancing execution coherence.

\begin{figure}[b!]
\vspace{-5pt}
\centering
\includegraphics[width=1.0\textwidth]{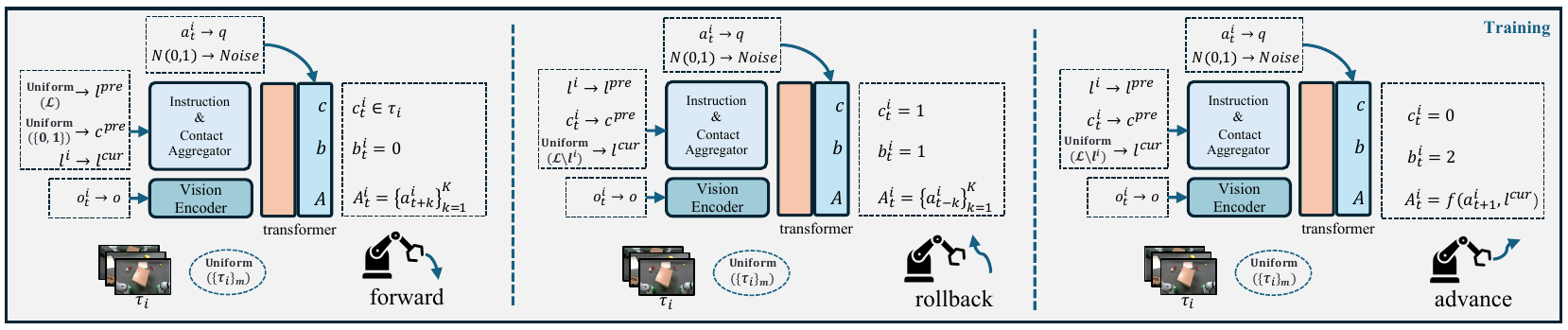}
\caption{
Illustration of the training pipelines for \texttt{forward}, \texttt{rollback}, and \texttt{advance} behaviors. Dynamic task transitions are achieved through policy modulation based on the predicted behavior mode, allowing the system to adapt to changing instructions and feedback.} 
\label{fig4}
\end{figure} 
\subsection{Training and Inference}
\label{sec:training_inference}

\subsubsection{Training with Behavior-Specific Conditioning}
Given labeled expert trajectories $\tau_i = \{(l^i | o^i_t, a^i_t, c^i_t, b^i_t)\}_{t=0}^{T}$, we design behavior-specific supervision to encourage the model to adapt its action generation based on both task instructions and physical interaction signals, as shown in Fig.~\ref{fig4}.
The behaviors are defined as follows:

\begin{itemize}
\item \textbf{$\texttt{forward}~(b_t = 0)$}: Predict the future action chunk $\{a^i_{t+1}, \dots, a^i_{t+K}\}$ under the matched instruction $l^{\text{cur}} = l^i$. Robustness to prior context is encouraged by randomly sampling different $l^{\text{pre}}, c^{\text{pre}}$ during training.

\item \textbf{$\texttt{rollback}~(b_t = 1)$}: Generate reversed actions $\{a^i_{t-1}, \dots, a^i_{t-K}\}$ when presented with a mismatched instruction $l^{\text{cur}} \neq l^i$ and active contact $c^i_t = 1$. This encourages the model to recover gracefully from semantic mismatches through physical feedback.

\item \textbf{$\texttt{advance}~(b_t = 2)$}: 
Interpolate linearly in joint space between the current action $a^i_{t}$ and a canonical start pose $a_0^{\text{normal}}$, defined as the mean of initial trajectories for instruction $l^{\text{cur}}$. This interpolation is represented by  $f(a^i_{t}, l^{\text{cur}})$, enabling rapid subtask switching under the assumption of no contact.
\end{itemize}

Action sequences are optimized using flow-matching loss~\cite{black2024pi_0}, which promotes smooth and dynamically feasible trajectory generation. Contact states $c^i_t$ and behavior modes $b^i_t$ are supervised via standard classification objectives, ensuring accurate perception of both environmental interaction and high-level intent.

\subsubsection{Inference with Conditional Switching}
During execution, the policy operates on three key inputs: the current visual observation $o_t$, a pair of instructions $(l^{\text{pre}}, l^{\text{cur}})$, and the previously propagated contact state $c^{\text{pre}}$. At each timestep, the model jointly predicts (i) the new contact state $c_t$ (to be used as $c^{\text{pre}}$ in the next step), (ii) the behavior mode $b_t \in \{0, 1, 2\}$, and (iii) the corresponding action chunk $A_t$. 

After each action chunk is executed, the model updates $l^{\text{pre}} \leftarrow l^{\text{cur}}$ and propagates the predicted contact state $c_t$ to the next step. New instructions can be issued at any timestep, upon which the model re-evaluates $b_t$ and dynamically adjusts behavior.

Depending on the predicted behavior mode, the system executes one of the following strategies: 
$\texttt{forward}~(b_t = 0)$: continue standard action prediction based on the current goal; 
$\texttt{rollback}~(b_t = 1)$: reverse previous actions to recover from errors under active contact; 
$\texttt{advance}~(b_t = 2)$: jump toward the new subgoal while updating the historical instruction as $l^{\text{pre}} \leftarrow l^{\text{cur}}$.

This conditional switching mechanism enables real-time, context-aware task transitions that couple semantic intent with physical interaction, supporting robust and adaptive task execution.

\section{Experiments}
\label{sec:exp}

\subsection{Task Protocol}
\label{sec:task-protocol}
Task switching occurs when the execution of \textit{Task A} is interrupted upon receipt of a new instruction for \textit{Task B}. Based on this, we perform two types of task switching experiments: pairwise evaluation and long sequence evaluation. To facilitate a comprehensive evaluation of the model's capabilities, in pairwise experiments, we send new instruction at different execution phases: early (pre-contact), mid (in-contact), and late (post-action). 
For long sequence experiments, we focus on mid-phase switching, as it offers a more rigorous evaluation of the method performance. A task switching is considered successful only if \textit{Task A} enters its designated execution phase without failure, and when new instruction is triggered, \textit{Task A} behaves as expected formulated in Section~\ref{sec:method}, and then \textit{Task B} is subsequently completed. Failure conditions include the inability to place an object (e.g., due to dropping) at the target position, non-compliance with a new instruction, or an unexpected execution halt. Each evaluation is repeated over 12 trials, with results averaged to ensure statistical reliability. 
 
Due to page constraints, extended experimental details, including setup, training procedures, model comparisons, and ablation results, are provided in Appendix (with its overview in Section~\ref{sec:appendix-overview}).

\begin{table}[t]
  \centering
  \caption{
      Average success rates (\%) on individual task (``No Switch'') and pairwise switching experiments in the LIBERO-Goal simulation.
  }
  \label{tab:sim-combined}
  \resizebox{0.7\textwidth}{!}{
    \begin{tabular}{lc|ccc}
    \toprule
    \textbf{Method} & \textbf{No Switch} & \textbf{Early Switch} & \textbf{Mid Switch} & \textbf{Late Switch} \\
    \midrule
    $\pi_0$~\cite{black2024pi_0}   & 92.3          & 40.7          & 8.3           & 10.2          \\
    OpenVLA-OFT~\cite{kim2025fine} & \textbf{98.0} & 40.6          & 11.1          & 13.0          \\
    \cmidrule(lr){1-5}
    \rowcolor{gray!15} 
    \ours & 93.0 & \textbf{93.5} & \textbf{50.9} & \textbf{68.7} \\
    \bottomrule
    \end{tabular}
  }
  \vspace{-5pt}
\end{table} 

\begin{table}[t]
  \centering
  \caption{
    Accumulated average success rate (\%) of long sequence switch performance on both simulation and real experiments.
    Note that while individual tasks are not restricted to the same one, 
    we ensure the use of distinct task pairs appeared in the long sequence.
  }
  \label{tab:long-seq}
  \resizebox{0.85\textwidth}{!}{
    \begin{tabular}{clccccc}
      \toprule
       & \multirow{2.5}{*}{\textbf{Method}} & \multicolumn{5}{c}{\textbf{Task Sequence Length}} \\
      \cmidrule(lr){3-7}
      & & \textbf{A$\rightarrow$B} & \textbf{A$\rightarrow$B$\rightarrow$C} &
        \textbf{A$\rightarrow$\dots$\rightarrow$D} & \textbf{A$\rightarrow$\dots$\rightarrow$E} & \textbf{A$\rightarrow$\dots$\rightarrow$F} \\
      \midrule
      \multirow{3}{*}{Simulation}
        & $\pi_0$~\cite{black2024pi_0}   & 0.0            & 0.0           & 0.0           & 0.0           & 0.0           \\
        & OpenVLA-OFT~\cite{kim2025fine} & 0.0            & 0.0           & 0.0           & 0.0           & 0.0           \\
        \rowcolor{gray!15} & \ours & \textbf{100.0} & \textbf{83.3} & \textbf{83.3} & \textbf{75.0} & \textbf{50.0} \\
      \cmidrule(lr){1-7}
      \multirow{4}{*}{Real} 
      & MT-ACT~\cite{bharadhwaj2023roboagent}    & 0.0           & 0.0           & 0.0           & 0.0           & 0.0           \\
      & Diffusion Policy~\cite{chi2023diffusion} & 0.0           & 0.0           & 0.0           & 0.0           & 0.0           \\
      & $\pi_0$~\cite{black2024pi_0}             & 0.0           & 0.0           & 0.0           & 0.0           & 0.0           \\
      \rowcolor{gray!15} & 
      \ours & \textbf{95.6} & \textbf{83.3} & \textbf{79.2} & \textbf{58.3} & \textbf{54.2} \\
      \bottomrule
    \end{tabular}%
    }
  \vspace{-5pt}
\end{table}

\subsection{Simulation Experiments}
\label{sec:simulated-experiments}

\textbf{Experiment Setup.} We conduct experiments on the LIBERO~\cite{liu2023libero} platform using the LIBERO-Goal suite, which emphasizes task diversity within a single table manipulation environment. 
We select 8 tasks for multi-task training, each with 50 expert trajectories. 
For pairwise evaluation, we sample 9 representative task combinations from this set, while for long sequence evaluation, 6 consecutive tasks are chosen to assess sequential task-switching performance. 
We predict absolute joint space to ensure proper \texttt{advance} calculation.

\textbf{Evaluation Results.} We compare \ours with $\pi_0$~\cite{black2024pi_0} and OpenVLA-OFT~\cite{kim2025fine}. The overall pairwise switching success rates are summarized in Table~\ref{tab:sim-combined}. we first compare single-task performance (``No Switch''), where \ours achieves comparable success rates to these baseline methods. In pairwise task switching experiments, we observe that the baseline methods achieve a success rate of approximately $40\%$ for early-phase switching, while demonstrating notably lower performance in mid- and late-phase switching scenarios ($\sim$$10\%$). In contrast, \ours exhibits robust performance across all switching phases, outperforming competing methods by a substantial margin. 
Notably, simulation tasks terminate with distinct end poses (without resetting to home position), introducing additional challenges for late-switch scenarios. Leveraging our \texttt{advance} mechanism, the robot autonomously transitions to subsequent tasks' starting positions, enabling natural task completion.
For more challenging mid-phase long-sequence evaluations, \ours demonstrates robustness by steadily handling 6 consecutive task-switching scenarios—achieving competitive performance where other methods fail in the initial A$\rightarrow$B transition, as shown in Table~\ref{tab:long-seq}.

\subsection{Real-World Experiments}
\label{sec:real-world-experiments}

\textbf{Experiment Setup.} Real-world experiments are conducted on two dual-armed Franka Emika Panda robot workstations, each executing four unique tasks. Data collection is performed via human-teleoperated demonstrations, covering foundational skills such as pushing, opening, and pick-and-place. Each task comprises 200 trajectories, recorded from two wrist-mounted cameras, one third-person RGB camera, and the robot's proprioceptive states. For evaluation, objects are randomly positioned within a predefined workspace region.

\begin{table}[t]
  \centering
  \renewcommand{\arraystretch}{1.25}
  \caption{
    Average success rates (\%) on real-world experiments.
  }
  \label{tab:realworld-combined}
  \begin{adjustbox}{max width=0.98\textwidth}
    \begin{tabular}{l|cccc|cccc}
      \toprule
      \multirow{2.5}{*}{\textbf{Method}}
        & \multicolumn{4}{c|}{\textbf{Workstation 1}} & \multicolumn{4}{c}{\textbf{Workstation 2}} \\
      \cmidrule(r){2-5}
      \cmidrule(l){6-9}
      & \textbf{No Switch} & \textbf{Early Switch} & \textbf{Mid Switch} & \textbf{Late Switch}
      & \textbf{No Switch} & \textbf{Early Switch} & \textbf{Mid Switch} & \textbf{Late Switch} \\
      \midrule
      MT-ACT~\cite{bharadhwaj2023roboagent}    
        & 56.3          & 0.0           & 0.0           & 4.9           & 50.0           & 0.0           & 0.0           & 0.0           \\
      DP~\cite{chi2023diffusion} 
        & 93.8          & 13.2          & 4.8           & 13.2          & 83.3           & 49.3          & 0.0           & 34.7          \\
      $\pi_0$~\cite{black2024pi_0}             
        & \textbf{97.9} & 50.1          & 0.0           & 32.0          & \textbf{100.0} & 75.0          & 0.0           & 64.6          \\
      \cmidrule(lr){1-9}
      \rowcolor{gray!15} \ours                                    
        & 95.9          & \textbf{99.3} & \textbf{95.1} & \textbf{75.0} & \textbf{100.0} & \textbf{95.1} & \textbf{96.5} & \textbf{94.4} \\
      \bottomrule
    \end{tabular}
  \end{adjustbox}
  \vspace{-5pt}  
\end{table}

\begin{figure}[t!]
  \centering
  \includegraphics[width=0.97\textwidth]{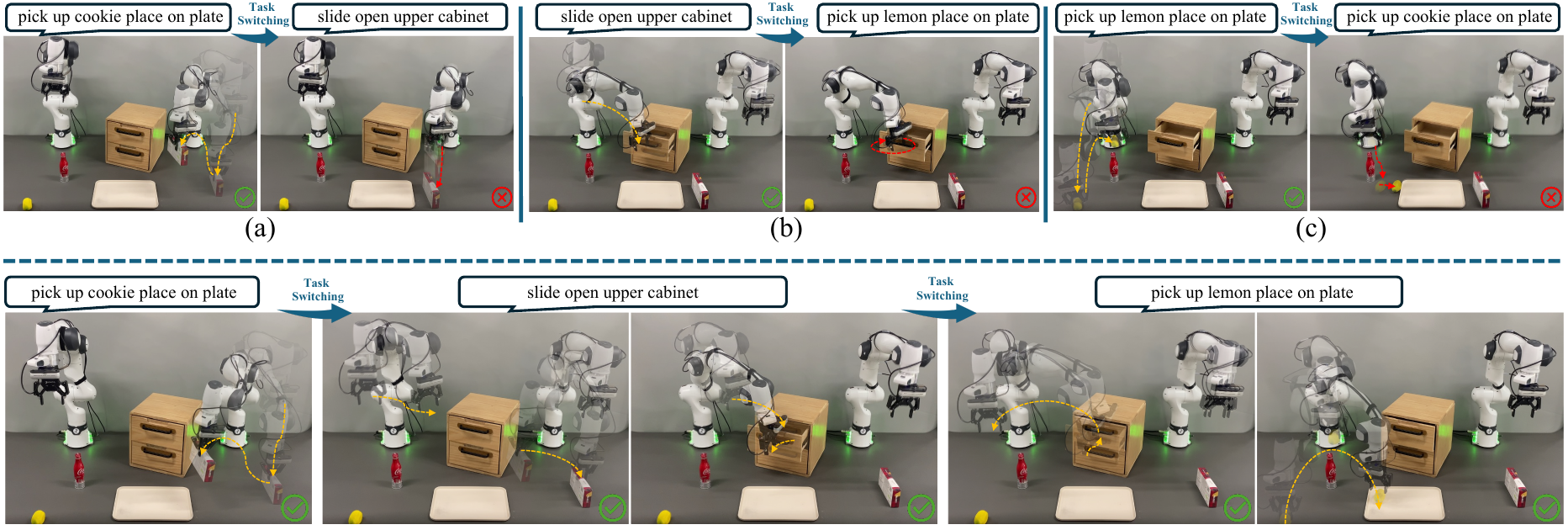}
  \caption{
    \textbf{Top:} 
      Performance of $\pi_0$~\cite{black2024pi_0} under pairwise task switching.
      (a), (b), and (c) each illustrate a unique task transition. Sudden switches during execution lead to erratic behaviors.
    \textbf{Bottom:} 
      \ours enables smooth and consistent and instruction-aligned task transitions.
  } 
  \label{fig5}
  \vspace{-5pt}
\end{figure} 
\begin{figure}[t]
  \centering
  \includegraphics[width=0.95\textwidth]{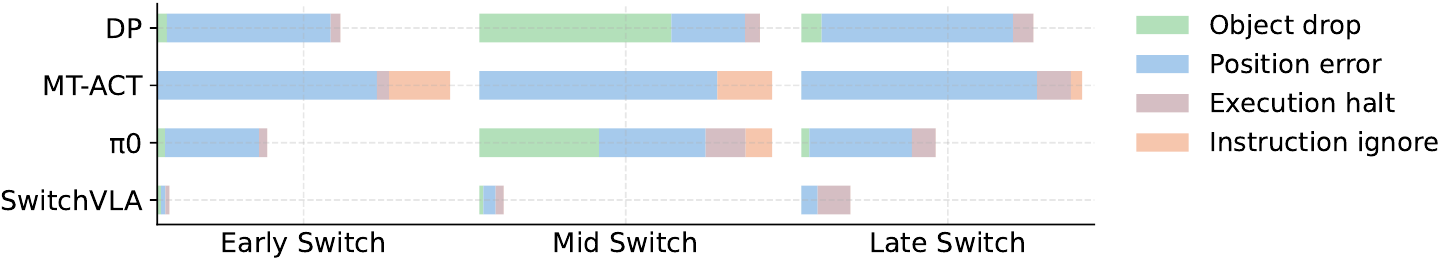}
  \caption{
    The proportion of failure causes during pairwise task switching in real-world experiments. Each bar's length represents the ratio of all failure occurrences across total trials.
  } 
  \label{fig6}
  \vspace{-5pt} 
\end{figure}

\textbf{Evaluation Results.} We evaluate \ours against three representative manipulation policies: MT-ACT~\cite{bharadhwaj2023roboagent}, Diffusion Policy (DP)~\cite{chi2023diffusion}, and $\pi_0$~\cite{black2024pi_0, allenzr2024openpi}. $\pi_0$ is a re-implementation based on the original paper. The results are summarized in Table~\ref{tab:realworld-combined}. Consistent with simulation results, \ours matches baseline performance in ``No Switch'' evaluations while maintaining robust success rates throughout all three switching phases. Similarly, long sequence experiments exhibit steady performance patterns as shown in Table~\ref{tab:long-seq}. We also present real-world experiment scenes in Fig.~\ref{fig5}, compared $\pi_0$ with \ours.
In real-world experiments, we observe distinct failure patterns across switching phases, as shown in Fig.~\ref{fig6}. During \textit{Early Switch}, failures often result from advancing to incorrect positions; MT-ACT additionally tends to ignore updated instructions. \textit{Mid Switch} is more challenging, with object drops and instruction neglect becoming common under active contact. In \textit{Late Switch}, execution halts and misplacements occur frequently due to incomplete subtask transitions. These patterns reflect the increasing complexity of dynamic switching. \ours effectively mitigates such failures through its instruction-conditioned, execution-aware policy.

\section{Conclusion}
We propose \textbf{\ours}, a unified execution-aware framework for Vision-Language-Action (VLA) robots.
By conditioning action generation on both task language and fine-grained execution signals, \ours enables seamless transitions across \texttt{forward}, \texttt{rollback}, and \texttt{advance} behaviors—without relying on modular planners or handcrafted switching logic.
Extensive experiments in both simulated and real-world environments demonstrate that \ours significantly outperforms prior VLA baselines in instruction adherence, task recovery, and switching robustness.
We position \ours as a more generalizable solution for instruction-conditioned control—capable of unifying diverse switching behaviors within a single policy framework.
We believe this offers a new perspective for designing interaction-aware VLA systems in open-ended environments. 
\section*{Limitations}
While SwitchVLA demonstrates strong performance in dynamic task switching, several limitations remain:
First, although our unified behavior-switching paradigm improves policy design, it relies on contact-based phase segmentation as a proxy for task progress. This may limit generalization to tasks involving abstract semantics or non-contact transitions. Incorporating richer temporal or semantic signals—such as intent prediction or temporal grounding—could address this issue.
Second, the current use of three discrete behavior modes (\texttt{forward}, \texttt{rollback}, \texttt{advance}) may constrain flexibility in scenarios involving ambiguous or drifting human intent. Learning continuous or adaptive behavior representations could provide greater expressiveness.
Finally, while our model is trained with weakly supervised paired demonstrations, adapting to noisy feedback, reinforcement refinement, or open-world settings remains an open challenge.

By introducing our approach and integrating it with high-level planning, we believe these limitations can be gradually addressed, extending the applicability of execution-aware VLA models in open-ended, interactive environments.



\bibliography{references}  

\clearpage
\appendix
\section*{\centering Appendix}

\section{Overview}
\label{sec:appendix-overview}
This appendix provides comprehensive technical details of \ours, organized into several key aspects. Section~\ref{sec:appendix-implementation-details} presents implementation specifics, including: (1) our dynamic condition-action data sampling strategy (Section~\ref{sec:appendix-data-sample}), (2) model architecture details (Section~\ref{sec:appendix-model-arch}), and (3) the contact state labeling methodology (Section~\ref{sec:appendix-contact-label}). Finally, we document the hyperparameter configurations used during training in Section~\ref{sec:appendix-hyper-params}.

We present additional experimental validation in Section~\ref{sec:appendix-additional-exps}, beginning with an ablation study of model components. We then compare \ours against baseline approaches that require specialized switching-behavior datasets, demonstrating our method's superior data efficiency.

Section~\ref{sec:appendix-eval-tasks} details our comprehensive task selection methodology. We begin by defining the fundamental single-task benchmarks, which serve as the basis for our evaluation. Building upon these atomic tasks, we then present:
\begin{itemize}
    \setlength{\itemindent}{10pt}
    \item Carefully constructed pairwise switching tasks to evaluate basic switching capabilities.
    \item Strategically sampled long sequence switching tasks to assess complex, multi-step transitions.
\end{itemize}

The complete experimental evaluation is documented in Section~\ref{sec:appendix-eval-results}, structured as:
\begin{itemize}
    \setlength{\itemindent}{10pt}
    \item Implementation setup and configuration details (Section~\ref{sec:appendix-exp-setup})
    \item Quantitative results for single-task and pairwise switching scenarios (Section~\ref{sec:appendix-exp-details}).
    \item Extended analysis of long-sequence task switching performance (Section~\ref{sec:appendix-exp-long-details}).
\end{itemize}
\section{Implementation Details}
\label{sec:appendix-implementation-details}
\subsection{Data Sampling Strategy}
\label{sec:appendix-data-sample}
To facilitate task switching learning without collecting additional demonstration data, we introduce a novel training paradigm that dynamically adapts action allocation according to both task instructions and physical interaction signals, shown in Section~\ref{sec:training_inference}. 
The detailed data sampling strategy for training \ours is presented in Algorithm~\ref{algo:data-sample}.
Importantly, while we demonstrate this framework with contact states and three distinct behaviors, the proposed sampling strategy is fundamentally extensible-other condition-action combinations can be implemented.

\subsection{Model Architecture Details}
\label{sec:appendix-model-arch}
\ours adopts Florence-2-base~\cite{xiao2023florence} architecture as its backbone with several key modifications. Following the original Florence-2-base design, we maintain a consistent token dimension of 768 throughout the model.

\textbf{Vision-Language-Contact Encoder.} Our Vision-Language-Contact (VLC) model consists of three key components. For visual processing, we adopt Florence-2-base's default image tokenizer, DaViT~\cite{ding2022davit}, which takes $3\times224\times224$ input images and produces 50 output tokens per image. For language processing, we utilize the default BART encoder~\cite{lewis2019bart}, processing all instructions to a fixed length of 20 tokens through padding. The contact state is encoded using a single-layer MLP, producing a 1-dimensional token. 
We construct the input representation by concatenating the previous instruction tokens ($l^{\text{pre}}$), current instruction tokens ($l^{\text{cur}}$), and previous contact state token ($c^{\text{pre}}$), forming a combined condition vector of length 41. This condition vector is then concatenated with the image tokens and processed through the transformer encoder, enabling multi-modal integration of visual, linguistic, and contact signals. 

\begin{algorithm}[t]
  \caption{Data Sampling Strategy in \ours Training}
  \label{algo:data-sample}
  \begin{algorithmic}[1]
    \renewcommand{\algorithmicrequire}{\textbf{Input:}}
    \Require Given one sampled observation $o_{t}$ from an episode at timestep $t$, full trajectory actions $A$, robot state $q_t$, robot action $a_{t}$ (with $q_t = a_t$), chunk size $K$, task label $l$, contact label $c_{t}$, all tasks $\mathcal{L}$, all contact state $\mathcal{C}=\{c_{0}, c_{1}\}$, task mode $\mathcal{M}=\{\text{current}, \text{previous}\}$, predefined probability to sample current task mode $P_{\text{cur}}$, predefined three behaviors $\mathcal{B}=\{b_{0}, b_{1}, b_{2}\}$, function that calculates the trajectory actions from current robot state to a specific task robot home state $f$.
    \State $L_{\text{mode}} \sim \text{Categorical}([P_{\text{cur}}, 1-P_{\text{cur}}])$ \Comment{Randomly assign obs to be cur or pre task}
    \If{$L_{\text{mode}} = \text{current}$} \Comment{\texttt{Forward} behavior}
        \State $l^{\text{cur}} \gets l$
        \State $l^{\text{pre}} \sim \text{Uniform}(\mathcal{L})$    \Comment{Sample a task that can be the same as the current task}
        \State $c^{\text{pre}} \sim \text{Uniform}(\{c_{0}, c_{1}\})$  
        \State $A_{t} \gets a_{t+1:t+K}$
        \State $b_{t} \gets b_{0}$
    \Else                               
        \State $l^{\text{cur}} \sim \text{Uniform}(\mathcal{L} \setminus l)$     \Comment{Sample a different task from the current task}
        \State $l^{\text{pre}} \gets l$

        \State $c^{\text{pre}} \gets c_{t}$
        \If{$c^{\text{pre}} = c_{0}$}                       \Comment{\texttt{Rollback} behavior}
            \State $A_{t} \gets a_{t-1:t-K}$                  \Comment{Revert actions}
            \State $b_{t} \gets b_{1}$
        \Else                                               \Comment{\texttt{Advance} behavior}
            \State $A_{t} \gets f(a_{t}, l^{\text{cur}})$   \Comment{Get trajectory from current pos to next task's home pos}
            \State $b_{t} \gets b_{2}$
        \EndIf
    \EndIf
    \State 

      \textbf{Return:} One data sample 
$ \gets \{\text{`observation': } o_{t}, \text{`action': } A_{t}, 
\text{`condition': } \{ \text{`task\_cur'}: l^{\text{cur}}, \text{`task\_pre': } l^{\text{pre}}, 
\text{`stage\_pre': } c^{\text{pre}}, \text{`behavior': } b_{t}\} \}$
  \end{algorithmic}
\end{algorithm} 

\textbf{Condition Execution Expert.} Our action expert builds upon Florence-2-base's transformer decoder, processing inputs from both the VLC encoder and a combined representation of aggregated conditions, robot proprioception, and action noise tokens (with chunk size $K$ for diffusion). The 41-dimensional condition tokens are first compressed to 5 dimensions via a single-layer MLP, while robot joint positions are encoded into a 1-dimensional token through another MLP. These components are concatenated with the action noise tokens, forming the expert's input (total dimension $5+1+K$).
The decoder outputs are processed through two separate 2-layer MLP projectors with SiLU activation: the first 5 tokens are decoded to predict contact state and behavior mode. For action generation, we employ diffusion flow-matching following \cite{allenzr2024openpi}. Our multi-task learning objective combines: (1) flow-matching MSE loss, (2) binary cross-entropy (BCE) loss for contact state prediction (weight $10^{-2}$), and (3) cross-entropy (CE) loss for behavior mode classification (weight $10^{-4}$).

\subsection{Contact State Label Acquisition}
\label{sec:appendix-contact-label}
Multiple approaches exist for acquiring robot-object contact information, including manual human annotation, force/torque sensing, vision-language models (VLMs), and gripper state monitoring. In our framework, we employ an automated pipeline for contact state labeling: First, we extract the current task's meta-data to identify (1) the active robotic arm and (2) relevant target objects. We then process the corresponding wrist-mounted camera image from the identified arm through VLMs to generate per-observation contact labels for entire trajectories.

For each trajectory, we generate contact state labels by sampling images at $1/3$ Hz and processing them through GPT-4o~\cite{gpt4o} to detect gripper-object contact. The visual analysis is guided by structured text prompts. To enhance prediction reliability, we integrate the gripper's opening status from proprioceptive observations, using this mechanical constraint to both refine contact predictions and eliminate false positives. An example of the prompt is shown below.

\vspace{15pt}

\begin{quote}
  \label{quote:prompt-example}
  \small
  \textbf{System Prompt:} \\
    \texttt{
      You are an expert in analyzing images from robotic arm grasping operations.
      You will receive an image from the hand-eye camera.
      Your task is to analyze this image according to the specific conditions provided by the user and determine if those conditions are met. 
      Return `1' if the conditions are satisfied and `0' if they are not.
      Do not provide any additional explanation.
    }
  
  \textbf{User Instruction:} \\
    \texttt{
    Analyze the provided image and determine whether:
    \begin{itemize}
      \item The robotic gripper has grasped/touched sandwich or its edge (return 1)
      \item The gripper is visibly closed (return 1)
      \item Neither condition is met (return 0)
    \end{itemize}
    }
\end{quote}

\vspace{2pt}

Finally, we apply 1D nearest-neighbor imputation to infer contact states for all observations in the trajectory. To further filter outliers, we employ a sliding window approach. Specifically, for a window size of $K$, we evaluate the contact results within the interval $[i - \lfloor K/2 \rfloor, i + \lfloor K/2 \rfloor]$ for each observation $O_i$. The contact state of $O_i$ is then assigned the \textit{mode} of the contact labels within this window, reducing noise and improving robustness.

\begin{table}[t]
  \centering
  \renewcommand{\arraystretch}{1.25}
  \caption{
    Contact state label accuracy (\%) for real-world workstation 1 and workstation 2.
    Refer to Sec~\ref{subsec:appendix-single-task-IDs} for task ID associations.
  }
  \label{tab:contact-accuracy}
  \vspace{7pt}
  \begin{adjustbox}{max width=0.65\textwidth}
    \begin{tabular}{cccc|cccc|c}
      \toprule
      \multicolumn{4}{c|}{\textbf{Workstation 1}} & \multicolumn{4}{c|}{\textbf{Workstation 2}} & \multirow{2.5}{*}{\textbf{Average}} \\
      \cmidrule(lr){1-4} \cmidrule(lr){5-8}
      \textbf{R1.1}      & \textbf{R1.2} & \textbf{R1.3} & \textbf{R1.4} & \textbf{R2.1} & \textbf{R2.2} & \textbf{R2.3} & \textbf{R2.4} & \\
      \midrule
      \rowcolor{gray!15} 91.7 & 88.0 & 80.8 & 60.6 & 79.5 & 84.2 & 73.9 & 86.7 & \textbf{80.7} \\
      \bottomrule
    \end{tabular}
  \end{adjustbox}
\end{table} 

The contact state information does not require high precision, as it serves primarily as coarse supervision for \ours to learn contact trends for relative prediction. For evaluation, we randomly sample $1,000$ observations (including images and contact state predictions) from each real-world task for human verification of labeling accuracy. As shown in Table~\ref{tab:contact-accuracy}, the average accuracy across all tasks reaches $80.7\%$, which is sufficient for \ours contact state prediction. During training, \ours contact state prediction accuracy on evaluation dataset is $89.3\%$, $98.6\%$, and $93.5\%$ respectively for workstation 1, workstation 2, and simulation. 

\subsection{Training Hyper-parameters}
\label{sec:appendix-hyper-params}
We train \ours with batch size of 512, using a constant learning rate scheduler which increases learning rate linearly between 0 and the initial learning rate with 5 warm-up steps. We train our models using the AdamW optimizer (with a learning rate of $5\times10^{-5}$ with weight decay of $10^{-4}$ and Exponential Moving Average (EMA). During training, we predict 100 future actions (chunk size) and use first 30 actions for temporal ensembling to allow smooth and accurate manipulation, following implementations in ACT~\cite{zhao2023learning}. In real-world training and inference, we use 3-RGB images, and predict 16-dimension actions (7-DoF$\times$2 \& 2 grippers). In simulation, we use 2-RGB images and predict 9-dimension actions (7-DoF \& 2 gripper tips). Both real-world and simulation share same other training settings. 

\begin{table}[t]
  \centering
  \caption{
      Average success rates (\%) of additional teleoperated switching data experiment on two pairwise switching tasks \textit{``pick up plate and place on plate''} \& \textit{``pick up basket sandwich and place on plate''}, and \textit{``pick up red gum and place on plate''} \& \textit{``push plate to customer''}.
  }
  \vspace{7pt}
  \label{tab:ablation-model}
  \resizebox{0.78\textwidth}{!}{
    \begin{tabular}{lccc}
    \toprule
    \textbf{Method} & \textbf{Early Switch} & \textbf{Mid Switch} & \textbf{Late Switch} \\
    \midrule
    \textit{SwitchVLA-base}   & 79.2          & 0.0          & 70.8           \\
    \textit{SwitchVLA-base$+$contact} & 87.5       & 95.8          & \textbf{100.0}          \\
    \cmidrule(lr){1-4}
    \rowcolor{gray!15} \textit{SwitchVLA-base$+$contact$+$behavior} & \textbf{91.7}       & \textbf{100.0}          & \textbf{100.0}          \\
    \bottomrule
    \end{tabular}
  }
\end{table}

\section{Additional Experiments}
\label{sec:appendix-additional-exps}
\textbf{Ablating Model Components.} We conduct an ablation study on \ours by examining two key components: contact state and behavior mode. Our real-world pairwise task-switching experiments reveal that removing these components reduces the model to a standard visual-language-action (VLA) imitation learning approach, which fails to determine appropriate task-switching timing during evaluation. 
As shown in Table~\ref{tab:ablation-model}, incorporating contact state information yields significant performance gains: early-switch accuracy improves from $79.2\%$ to $87.5\%$, mid-switch from $0.0\%$ to $95.8\%$, and late-switch from $70.8\%$ to $100\%$. The addition of behavior mode modeling as an auxiliary loss further enhances performance, achieving $91.7\%$ (early), $100.0\%$ (mid), and $100.0\%$ (late) success rates. These results demonstrate that both components are crucial for robust task-switching performance.

\begin{table}[t]
  \centering
  \caption{
      Average success rates (\%) of additional teleoperated switching data experiment on pairwise switching task \textit{``pick up plate and place on plate''} \& \textit{``pick up basket sandwich and place on plate''}.
  }
  \vspace{7pt}
  \label{tab:ablation-data}
  \resizebox{0.7\textwidth}{!}{
    \begin{tabular}{lccc}
    \toprule
    \textbf{Method} & \textbf{Early Switch} & \textbf{Mid Switch} & \textbf{Late Switch} \\
    \midrule
    \textit{SwitchVLA-base}   & 75.0          & 0.0          & 66.7           \\
    \textit{SwitchVLA-base-add-data} & \textbf{83.3}       & 91.7          & \textbf{100.0}          \\
    \cmidrule(lr){1-4}
    \rowcolor{gray!15} \ours         & \textbf{83.3} & \textbf{100.0} & \textbf{100.0} \\
    \bottomrule
    \end{tabular}
  }
\end{table}

\textbf{Comparison with additional behavior specific data} To demonstrate the data efficiency and robustness of \ours, we additionally collect teleoperated task switching data and train the \textit{\ours-base}. Specifically, we focus on a pairwise switch task—\textit{``Pick up plate place on plate''} switch to \textit{``Pick up basket sandwich place on plate''}. We collect $50$ early switch data, $100$ mid switch data, and $50$ late switch data, combined with original data for both tasks ($200$ each) to train \textit{\ours-base-add-data}. As shown in Table~\ref{tab:ablation-data}, we compare \textit{\ours-base}, \textit{\ours-base-add-data}, and \ours. The results demonstrate that additional training data improves the baseline model's performance. Our analysis suggests that with sufficient additional behavior-specific data, the baseline model could achieve performance comparable to or even surpassing \ours. However, such an approach demands substantial data curation efforts, which are neither cost-effective nor scalable. In contrast, \ours effectively addresses these challenges without relying on extensive additional data, demonstrating superior efficiency and practicality.

\section{Evaluation Tasks}
\label{sec:appendix-eval-tasks}
\begin{figure}[t]
  \centering
  \includegraphics[width=1.0\textwidth,height=0.31\textwidth]{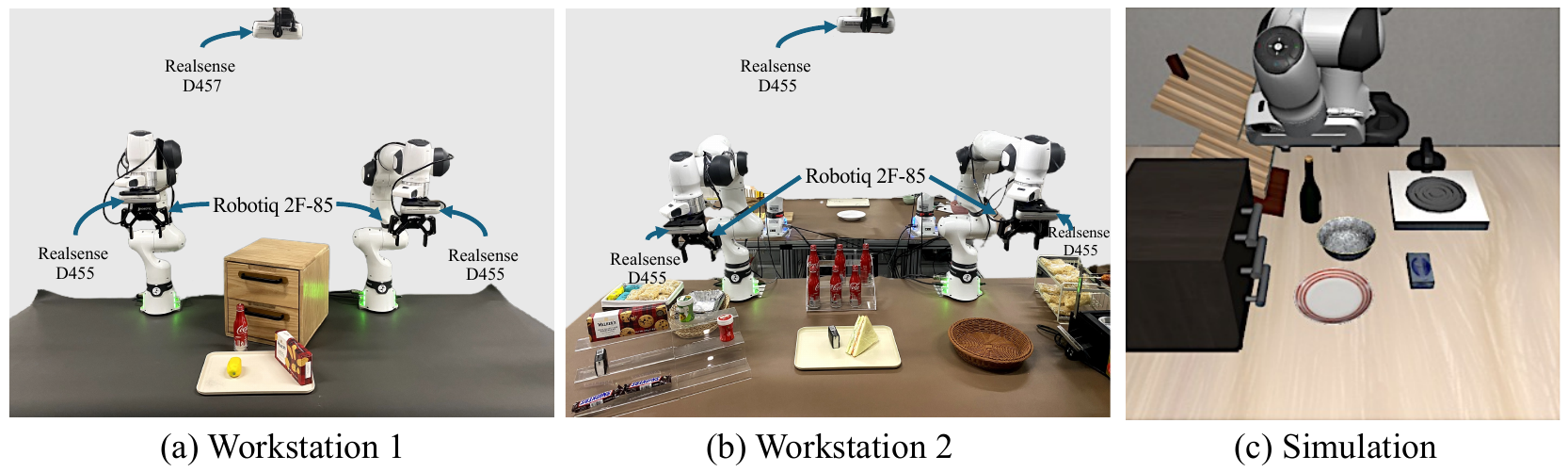}
  \caption{
    (a) Dual-armed Franka workstation 1 setup.
    (b) Dual-armed Franka workstation 2 setup.
    (c) LIBERO-Goal simulation setup. 
  }
  \label{fig:all-setups}
\end{figure}

In this section, we provide more details of real-world and simulation task definitions and evaluation results. The overview of experiment setups for real-world and simulation is shown in Fig.~\ref{fig:all-setups}. 

\subsection{Single Tasks} 
\label{subsec:appendix-single-task-IDs}
In workstation 1 and workstation 2, we setup the scenes and create $4$ unique single tasks for each workstation, based on the single tasks, we combine two tasks and result in $12$ pairwise switching tasks for each workstation. For LIBERO-Goal simulator, we select $8$ unique single tasks and create $9$ pairwise tasks. We describe all tasks below.

\subsubsection{Real Franka Workstation 1}
\textbf{Task R1.1: pick up lemon place on plate.} A basic pick-and-place task where the robot’s goal is to grasp a lemon from a specified location on table and place it onto the plate, evaluating its ability to handle small objects with precision.  \\
\textbf{Task R1.2: pick up coke bottle place on plate.} This is a more challenging task where the robot needs to identify and lift a coke bottle from table and place it on a plate, assessing its capability to manipulate the small bottle cap with more precision.  \\
\textbf{Task R1.3: pick up cookie place on plate}. This task involves carefully picking up a fragile cookie box without breaking it and placing it on a plate, testing the robot gripper's precise manipulation skills.  \\
\textbf{Task R1.4: slide open upper cabinet.} The robot’s target is to interact with an articulated cabinet door by sliding it open, evaluating its ability to handle constrained motion in a semi-structured environment.

\subsubsection{Real Franka Workstation 2}
\textbf{Task R2.1: pick up red gum place on plate.} The robot must locate and grasp a small red gum bottle from among multiple objects and place it on the plate, testing object discrimination and precise manipulation.  \\
\textbf{Task R2.2: pick up tissue place on plate.} A challenging task where the robot’s objective is to pick up a pack of tissue and place it accurately on the plate, assessing its ability to handle deformable and flexible materials.  \\
\textbf{Task R2.3: pick up sandwich place on plate.} The robot’s goal is to grasp a sandwich without deforming it and place it on the plate.  \\
\textbf{Task R2.4: push plate to customer.} The robot needs to push a plate forward to simulate serving, testing its ability to perform controlled planar manipulation while maintaining object stability.

\subsubsection{Simulation} 
We show simple description of tasks in LIBERO-Goal table manipulation environment here, refer to original paper~\cite{liu2023libero} for more details. \\
\textbf{Task S1: put the bowl on the plate.} A nested placement task where the robot must position a bowl securely on a plate, assessing spatial awareness and stability awareness.  \\
\textbf{Task S2: put the cream cheese in the bowl.} The robot’s objective is to pick up cream cheese and place into a bowl, evaluating its ability to perform precise manipulation.  \\
\textbf{Task S3: turn on the stove.} The robot needs to interact with a stove knob to turn it on, testing its capability to operate mechanical controls with rotational motion.  \\
\textbf{Task S4: put the bowl on the stove.} The robot’s target is to place a bowl onto a stove surface, requiring careful positioning and assess balanced manipulation.  \\
\textbf{Task S5: put the wine bottle on top of the cabinet.} The robot must lift and place a wine bottle onto a high cabinet shelf, testing its ability to perform tasks requiring height adjustment and careful object placement.  \\
\textbf{Task S6: push the plate to the front of the stove.} The robot’s goal is to slide a plate forward near the stove, evaluating controlled planar movement and spatial awareness in a kitchen environment.  \\
\textbf{Task S7: put the wine bottle on the rack.} The robot needs to place the wine bottle into a rack, assessing precision in constrained placement tasks with orientation sensitivity.  \\
\textbf{Task S8: put the bowl on top of the cabinet.} The robot’s objective is to lift and position a bowl on an elevated cabinet surface, testing its ability to handle objects and maintaining stability.  

\begin{table}[t]
  \centering
  \caption{Selected pairwise switching tasks for real-world and simulation evaluation.}
  \vspace{7pt}
  \label{tab:pairwise-task}
  \large 
  \resizebox{1\textwidth}{!}{
    \begin{tabular}{*{2}{w{c}{3cm}} | *{2}{w{c}{3cm}} | *{2}{w{c}{3cm}}}
    \toprule
    \multicolumn{2}{c|}{\textbf{Workstation 1}} & \multicolumn{2}{c|}{\textbf{Workstation 2}} & \multicolumn{2}{c}{\textbf{Simulation}} \\
    \cmidrule(lr){1-2} \cmidrule(lr){3-4} \cmidrule(lr){5-6}
    \textbf{Pairwise Task ID} & \textbf{Task Pairs} & \textbf{Pairwise Task ID} & \textbf{Task Pairs} & \textbf{Pairwise Task ID} & \textbf{Task Pairs} \\
    \midrule
    PR1.1 & R1.1$\rightarrow$R1.2 & PR2.1 & R2.1$\rightarrow$R2.2 & PS1 & S1$\rightarrow$S3 \\
    PR1.2 & R1.1$\rightarrow$R1.3 & PR2.2 & R2.2$\rightarrow$R2.1 & PS2 & S4$\rightarrow$S1 \\
    PR1.3 & R1.2$\rightarrow$R1.1 & PR2.3 & R2.1$\rightarrow$R2.3 & PS3 & S5$\rightarrow$S6 \\
    PR1.4 & R1.2$\rightarrow$R1.3 & PR2.4 & R2.3$\rightarrow$R2.1 & PS4 & S2$\rightarrow$S6 \\
    PR1.5 & R1.3$\rightarrow$R1.1 & PR2.5 & R2.1$\rightarrow$R2.4 & PS5 & S3$\rightarrow$S5 \\
    PR1.6 & R1.3$\rightarrow$R1.2 & PR2.6 & R2.4$\rightarrow$R2.1 & PS6 & S5$\rightarrow$S3 \\
    PR1.7 & R1.4$\rightarrow$R1.1 & PR2.7 & R2.2$\rightarrow$R2.3 & PS7 & S3$\rightarrow$S2 \\
    PR1.8 & R1.1$\rightarrow$R1.4 & PR2.8 & R2.3$\rightarrow$R2.2 & PS8 & S8$\rightarrow$S3 \\
    PR1.9 & R1.4$\rightarrow$R1.2 & PR2.9 & R2.2$\rightarrow$R2.4 & PS9 & S5$\rightarrow$S6 \\
    PR1.10 & R1.2$\rightarrow$R1.4 & PR2.10 & R2.4$\rightarrow$R2.2 &  & \\
    PR1.11 & R1.4$\rightarrow$R1.3 & PR2.11 & R2.3$\rightarrow$R2.4 &  &  \\
    PR1.12 & R1.3$\rightarrow$R1.4 & PR2.12 & R2.4$\rightarrow$R2.3 &  &  \\
    \bottomrule
    \end{tabular}
  }
\end{table}

\subsection{Pairwise Switching Tasks}
\label{subsec:appendix-pairwise-task-IDs}
We pair single tasks to form our pairwise switching tasks for real-world and simulation. 
We use all permutations for 4 tasks taken 2, \textit{i.e.} \( {}^4P_2 = 12 \) for two each real-world workstation. 
In LIBERO-Goal simulation, we sample 9 pairs. Full details are shown in Table~\ref{tab:pairwise-task}.

\begin{table}[t]
  \centering
  \caption{Selected long sequence tasks for real-world and simulation evaluation.}
  \label{tab:long-seq-task}
  \vspace{7pt}
  {\footnotesize
  \resizebox{0.9\textwidth}{!}{
    \begin{tabular}{c|>{\centering\arraybackslash}p{3cm}|>{\centering\arraybackslash}p{7cm}}
    \toprule
     & \multicolumn{1}{c|}{\textbf{Long-Seq Task ID}} & \multicolumn{1}{c}{\textbf{Task Sequence}} \\
    \midrule
    \textbf{Workstation 1} & LR1 & R1.1$\rightarrow$R1.3$\rightarrow$R1.2$\rightarrow$R1.1$\rightarrow$R1.2$\rightarrow$R1.4 \\
    \cmidrule(lr){1-3}
    \textbf{Workstation 2} & LR2 & R2.1$\rightarrow$R2.3$\rightarrow$R2.4$\rightarrow$R2.2$\rightarrow$R2.3$\rightarrow$R2.1 \\
    \cmidrule(lr){1-3}
    \multirow{4}{*}{\textbf{Simulation}} & LS1 & S2$\rightarrow$S5$\rightarrow$S4$\rightarrow$S1$\rightarrow$S8$\rightarrow$S3 \\
    \cmidrule(r){2-3}
    & LS2 & S1$\rightarrow$S3$\rightarrow$S2$\rightarrow$S8$\rightarrow$S3$\rightarrow$S6 \\
    \cmidrule(r){2-3}
    & LS3 & S8$\rightarrow$S3$\rightarrow$S1$\rightarrow$S3$\rightarrow$S7$\rightarrow$S4 \\
    \bottomrule
    \end{tabular}
  }}
\end{table}

\subsection{Long Sequence Switching Tasks}
\label{subsec:appendix-long-task-IDs}
In Section~\ref{sec:exp}, we show long sequence switching task performance in simulation environment for task LS1, the task selection is shown in Table~\ref{tab:long-seq-task}, all other methods show $0.0\%$ success rates, as they by chance fail at the beginning S2$\rightarrow$S5 pairwise tasks. 
We samples two more long sequence—LS2 and LS3 tasks for comparison.

\section{Evaluation Results}
\label{sec:appendix-eval-results}
In this section, we present experimental environment setup, as illustrated in Fig.~\ref{fig:all-setups}, and evaluation details, including detailed results for polices on single and pairwise switching.

\subsection{Experiment Setup}
\label{sec:appendix-exp-setup}
\textbf{Workstations.} We use 7-DoF Franka Emika plus RobotiQ 2F-85 grippers setup with three Intel RealSense D455 or D457 cameras, mounted at left arm wrist, right arm wrist and third-person top view. The data collection and inference frequency is $15Hz$. All models employ $224\times224$ image size for both training and inference. During inference, all models are deployed on a single NVIDIA RTX 4090 GPU.

\begin{table}[t]
  \centering
  \caption{Simulation joint position controller parameters.}
  \label{tab:sim-params}
  \vspace{7pt}
  {\footnotesize 
  \resizebox{0.5\textwidth}{!}{ 
    \begin{tabular}{>{\centering\arraybackslash}p{3cm}|>{\centering\arraybackslash}p{3cm}}
    \toprule
    \textbf{Parameters} & \textbf{Value} \\
    \midrule
    output\_min & $1.0$ \\
    \cmidrule(lr){1-2}
    output\_max & $1.0$ \\
    \cmidrule(lr){1-2}
    kp & $15,000.0$ \\
    \cmidrule(lr){1-2}
    ramp\_ratio & $1.0$ \\
    \bottomrule
    \end{tabular}
  }}
\end{table}

\textbf{Simulation.} In LIBERO-Goal, we employ both wrist camera and third-person camera for training and inference. The image size is $128\times128$. Each task contains 50 human-teleoperated demonstrations.
We keep the same setting of each LIBERO-Goal task using its corresponding BDDL file, while using \textit{JOINT\_POSITION} control and changing some parameters in Robosuite~\cite{ahn2022can} to keep proper joint position inference on simulation (default inference uses end effector \textit{OSC\_POSE} control). The configurations are shown in Table~\ref{tab:sim-params}. 
Same to real-world setup, during inference, all models are deployed on a single NVIDIA RTX 4090 GPU.

\begin{table}[t]
  \centering
  \renewcommand{\arraystretch}{1.25}
  \caption{
    Average success rates (\%) on all single task experiments. We run 50 trials for each simulation evaluation, and 12 trials for real-world evaluation.
  }
  \label{tab:appendix-single-results}
  \begin{adjustbox}{max width=0.8\textwidth}
    \begin{tabular}{l|cccccccc|c}
      \toprule
      \multirow{2.5}{*}{\textbf{Method}}
        & \multicolumn{4}{c|}{\textbf{Workstation 1}} & \multicolumn{4}{c|}{\textbf{Workstation 2}} & \multirow{2.5}{*}{\textbf{Average (\%)}} \\
      \cmidrule(lr){2-5}
      \cmidrule(lr){6-9}
      & \textbf{R1.1} & \textbf{R1.2} & \textbf{R1.3} & \multicolumn{1}{c|}{\textbf{R1.4}}
      & \textbf{R2.1} & \textbf{R2.2} & \textbf{R2.3} & \textbf{R2.4} \\
      \midrule
      MT-ACT~\cite{bharadhwaj2023roboagent}    
        & 75.0 & 83.3 & 66.7 & \multicolumn{1}{c|}{0.0} & 100.0 & 16.7 & 83.3 & 0.0 & 53.1 \\
      DP~\cite{chi2023diffusion} 
        & 83.3 & 91.7 & 100.0 & \multicolumn{1}{c|}{100.0} & 100.0 & 91.7 & 83.3 & 58.3 & 88.5 \\
      $\pi_0$~\cite{black2024pi_0}             
         & 100.0 & 91.7 & 100.0 & \multicolumn{1}{c|}{100.0} & 100.0 & 100.0 & 100.0 & 100.0 & 99.0 \\
      \rowcolor{gray!15} \ours     
         & 100.0 & 91.7 & 91.7 & \multicolumn{1}{c|}{100.0} & 100.0 & 100.0 & 100.0 & 100.0 & 97.9 \\
      \midrule
      \multirow{2.5}{*}{\textbf{Method}} & \multicolumn{8}{c|}{\textbf{Simulation}} & \multirow{2.5}{*}{\textbf{Average (\%)}} \\
      \cmidrule(lr){2-9}
      & \textbf{S1} & \textbf{S2} & \textbf{S3} & \textbf{S4} & \textbf{S5} & \textbf{S6} & \textbf{S7} & \textbf{S8} \\
      \midrule
      OpenVLA-OFT~\cite{kim2025fine} & 100.0 & 100.0 & 100.0 & 94.0 & 94.0 & 100.0 & 98.0 & 98.0 & 98.0 \\
      $\pi_0$~\cite{black2024pi_0}   & 98.0  & 96.0  & 100.0 & 96.0 & 98.0 & 90.0  & 76.0 & 88.0 & 92.3 \\
      \rowcolor{gray!15} \ours 
                  & 100.0 & 62.0  & 100.0 & 100.0 & 100.0 & 82.0 & 100.0 & 100.0 & 93.0 \\
      \bottomrule
    \end{tabular}
  \end{adjustbox}
\end{table} 
\begin{table}[t]
  \centering
  \caption{Average success rate (\%) over 12 trials on all pairwise task switching experiments.}
  \label{tab:appendix-pair-results}
  \vspace{7pt}
  {\Large 
  \resizebox{1\textwidth}{!}{ 
    \begin{tabular}{
        >{\centering\arraybackslash}p{3.5cm}
        >{\centering\arraybackslash}p{1.5cm}
        >{\centering\arraybackslash}p{2cm}
        >{\centering\arraybackslash}p{2cm}
        >{\centering\arraybackslash}p{2cm}
        >{\centering\arraybackslash}p{2cm}
        >{\centering\arraybackslash}p{2cm}
        >{\centering\arraybackslash}p{2cm}
        >{\centering\arraybackslash}p{2cm}
        >{\centering\arraybackslash}p{2cm}
        >{\centering\arraybackslash}p{2cm}
        >{\centering\arraybackslash}p{2cm}
        >{\centering\arraybackslash}p{2cm}
        >{\centering\arraybackslash}p{2cm}|
        >{\centering\arraybackslash}p{2cm} 
      }
      \toprule
      \multirow{2}{*}{\textbf{Method}} & \textbf{Switch Stage} & 
      \multicolumn{12}{c}{\multirow{2}{*}{\textbf{Pairwise Switching Experiment Task IDs}}} & \textbf{Average (\%)} \\
      \midrule
      & & \multicolumn{12}{c}{\textbf{Workstation 1}} & \\
      \cmidrule(lr){3-14}
      & & \textbf{PR1.1} & \textbf{PR1.2} & \textbf{PR1.3} & \textbf{PR1.4} & \textbf{PR1.5} & \textbf{PR1.6} & \textbf{PR1.7} & \textbf{PR1.8} & \textbf{PR1.9} & \textbf{PR1.10} & \textbf{PR1.11} & \multicolumn{1}{c}{\textbf{PR1.12}} &  \\
      \cmidrule(lr){1-15}
      MT-ACT & \multirow{4}{*}{Early} & 0.0 & 0.0 & 0.0 & 0.0 & 0.0 & 0.0 & 0.0 & 0.0 & 0.0 & 0.0 & 0.0 & 0.0 & 0.0 \\
      DP & & 0.0 & 58.3 & 0.0 & 41.7 & 0.0 & 0.0 & 58.3 & 0.0 & 0.0 & 0.0 & 0.0 & 0.0 & 13.2 \\
      $\pi_0$ & & 41.7 & 16.7 & 16.7 & 41.7 & 58.3 & 75.0 & 0.0 & 91.7 & 16.7 & 83.3 & 58.3 & 100.0 & 50.1 \\
       \rowcolor{gray!15} \ours & & 100.0 & 100.0 & 100.0 & 91.7 & 100.0 & 100.0 & 100.0 & 100.0 & 100.0 & 100.0 & 100.0 & 100.0 & \textbf{99.3} \\
      \cmidrule(lr){1-15}
      MT-ACT & \multirow{4}{*}{Mid} & 0.0 & 0.0 & 0.0 & 0.0 & 0.0 & 0.0 & 0.0 & 0.0 & 0.0 & 0.0 & 0.0 & 0.0 & 0.0 \\
      DP & & 0.0 & 0.0 & 0.0 & 0.0 & 0.0 & 0.0 & 0.0 & 0.0 & 0.0 & 0.0 & 58.3 & 0.0 & 4.8 \\
      $\pi_0$ & & 0.0 & 0.0 & 0.0 & 0.0 & 0.0 & 0.0 & 0.0 & 0.0 & 0.0 & 0.0 & 0.0 & 0.0 & 0.0 \\
       \rowcolor{gray!15} \ours & & 100.0 & 83.3 & 100.0 & 100.0 & 100.0 & 100.0 & 91.7 & 83.3 & 100.0 & 91.7 & 100.0 & 91.7 & \textbf{95.1} \\
      \cmidrule(lr){1-15}
      MT-ACT & \multirow{4}{*}{Late} & 41.7 & 0.0 & 16.7 & 0.0 & 0.0 & 0.0 & 0.0 & 0.0 & 0.0 & 0.0 & 0.0 & 0.0 & 4.9 \\
      DP & & 50.0 & 0.0 & 25.0 & 0.0 & 0.0 & 0.0 & 0.0 & 41.7 & 0.0 & 0.0 & 41.7 & 0.0 & 13.2 \\
      $\pi_0$ & & 41.7 & 41.7 & 33.3 & 50.0 & 66.7 & 50.0 & 0.0 & 0.0 & 0.0 & 41.7 & 0.0 & 58.3 & 32.0 \\
       \rowcolor{gray!15} \ours & & 50.0 & 91.7 & 100.0 & 100.0 & 91.7 & 100.0 & 0.0 & 83.3 & 0.0 & 83.3 & 100.0 & 100 & \textbf{75.0} \\
      \midrule
      & & \multicolumn{12}{c}{\textbf{Workstation 2}} & \\
      \cmidrule(lr){3-14}
      & & \textbf{PR2.1} & \textbf{PR2.2} & \textbf{PR2.3} & \textbf{PR2.4} & \textbf{PR2.5} & \textbf{PR2.6} & \textbf{PR2.7} & \textbf{PR2.8} & \textbf{PR2.9} & \textbf{PR2.10} & \textbf{PR2.11} & \multicolumn{1}{c}{\textbf{PR2.12}} & \\
      \cmidrule(lr){1-15}
      MT-ACT & \multirow{4}{*}{Early} & 0.0 & 0.0 & 0.0 & 0.0 & 0.0 & 0.0 & 0.0 & 0.0 & 0.0 & 0.0 & 0.0 & 0.0 & 0.0 \\
      DP & & 33.3 & 0.0 & 100.0 & 58.3 & 0.0 & 0.0 & 100.0 & 100.0 & 0.0 & 100.0 & 0.0 & 100.0 & 49.3 \\
      $\pi_0$ & & 100.0 & 0.0 & 100.0 & 0.0 & 100.0 & 0.0 & 100.0 & 100.0 & 100.0 & 100.0 & 100.0 & 100.0 & 75.0 \\
       \rowcolor{gray!15} \ours & & 100.0 & 100.0 & 58.3 & 100.0 & 100.0 & 100.0 & 83.3 & 100.0 & 100.0 & 100.0 & 100.0 & 100.0 & \textbf{95.1} \\
      \cmidrule(lr){1-15}
      MT-ACT & \multirow{4}{*}{Mid} & 0.0 & 0.0 & 0.0 & 0.0 & 0.0 & 0.0 & 0.0 & 0.0 & 0.0 & 0.0 & 0.0 & 0.0 & 0.0 \\
      DP & & 0.0 & 0.0 & 0.0 & 0.0 & 0.0 & 0.0 & 0.0 & 0.0 & 0.0 & 0.0 & 0.0 & 0.0 & 0.0 \\
      $\pi_0$ & & 0.0 & 0.0 & 0.0 & 0.0 & 0.0 & 0.0 & 0.0 & 0.0 & 0.0 & 0.0 & 0.0 & 0.0 & 0.0 \\
       \rowcolor{gray!15} \ours & & 100.0 & 100.0 & 83.3 & 100.0 & 100.0 & 100.0 & 100.0 & 83.3 & 100.0 & 91.7 & 100.0 & 100.0 & \textbf{96.5} \\
      \cmidrule(lr){1-15}
      MT-ACT & \multirow{4}{*}{Late} & 0.0 & 0.0 & 0.0 & 0.0 & 0.0 & 0.0 & 0.0 & 0.0 & 0.0 & 0.0 & 0.0 & 0.0 & 0.0 \\
      DP & & 58.3 & 0.0 & 0.0 & 66.7 & 0.0 & 0.0 & 83.3 & 100.0 & 66.7 & 0.0 & 41.7 & 0.0 & 34.7 \\
      $\pi_0$ & & 100.0 & 100.0 & 100.0 & 100.0 & 100.0 & 0.0 & 0.0 & 100.0 & 100.0 & 0.0 & 75.0 & 0.0 & 64.6 \\
       \rowcolor{gray!15} \ours & & 100.0 & 100.0 & 100.0 & 100.0 & 100.0 & 83.3 & 100.0 & 100.0 & 100.0 & 100.0 & 50.0 & 100.0 & \textbf{94.4} \\
      \midrule
      & & \multicolumn{9}{c}{\textbf{Simulation}} & \multicolumn{4}{c}{} \\
      \cmidrule(lr){3-11}
      & & \textbf{PS1} & \textbf{PS2} & \textbf{PS3} & \textbf{PS4} & \textbf{PS5} & \textbf{PS6} & \textbf{PS7} & \textbf{PS8} & \textbf{PS9} & \multicolumn{4}{c}{} \\
      \cmidrule(lr){1-15}
      OpenVLA-OFT & \multirow{3}{*}{Early} & 91.7 & 33.3 & 25.0 & 66.7 & 33.3 & 33.3 & 33.3 & 25.0 & 25.0 & & & & 40.6 \\
      $\pi_0$ & & 25.0 & 8.3 & 33.3 & 83.3 & 33.3 & 66.7 & 41.7 & 41.7 & 33.3 & & & & 40.7 \\
       \rowcolor{gray!15} \ours & & 100.0 & 91.7 & 75.0 & 100.0 & 91.7 & 100.0 & 83.3 & 100.0 & 100.0 & & & & \textbf{93.5} \\
      \cmidrule(lr){1-15}
      OpenVLA-OFT & \multirow{3}{*}{Mid} & 41.7 & 0.0 & 0.0 & 0.0 & 0.0 & 0.0 & 0.0 & 58.3 & 0.0 & & & & 11.1 \\
      $\pi_0$ & & 0.0 & 0.0 & 0.0 & 58.3 & 0.0 & 0.0 & 16.7 & 0.0 & 0.0 & & & & 8.3 \\
       \rowcolor{gray!15} \ours & & 100.0 & 83.3 & 33.3 & 8.3 & 33.3 & 91.7 & 41.7 & 66.7 & 0.0 & & & & \textbf{50.9} \\
      \cmidrule(lr){1-15}
      OpenVLA-OFT & \multirow{3}{*}{Late} & 0.0 & 100.0 & 0.0 & 16.7 & 0.0 & 0.0 & 0.0 & 0.0 & 0.0 & & & & 13.0 \\
      $\pi_0$ & & 0.0 & 0.0 & 0.0 & 16.7 & 0.0 & 8.3 & 0.0 & 66.7 & 0.0 & & & & 10.2 \\
       \rowcolor{gray!15} \ours & & 100.0 & 41.7 & 41.7 & 66.7 & 25.0 & 100.0 & 41.7 & 100.0 & 100.0 & & & & \textbf{68.7} \\
      \bottomrule
    \end{tabular}
  }}
\end{table} 

\subsection{Evaluation details}
\label{sec:appendix-exp-details}
We evaluate \ours against three prior manipulation policies and selected the more representative real-robot works: MT-ACT~\cite{bharadhwaj2023roboagent}, Diffusion Policy (DP)~\cite{chi2023diffusion}, and $\pi_0$~\cite{black2024pi_0,allenzr2024openpi}. $\pi_0$ is a re-implementation based on the original paper.
For simulation, we compare with state-of-the-art (SOTA) VLAs—$\pi_0$~\cite{black2024pi_0} and OpenVLA-OFT~\cite{kim2025fine} on simulation.
We show all individual task results in Table~\ref{tab:appendix-single-results}, and pairwise results in Table~\ref{tab:appendix-pair-results}. For single-task performance (``No Switch''), \ours demonstrates comparable performance to selected baseline methods. In pairwise task switching experiments, we observe that the baseline methods achieve higher for early- and late-switching, while showing notably lower performance in mid-switching tasks. In contrast, \ours exhibits robust performance across all switching phases, outperforming competing methods by a substantial margin.

\begin{table}[t]
  \centering
  \caption{
    Accumulated average success rate (\%) of long sequence switch performance on simulation experiments.
    Note that while individual tasks are not restricted to the same one, we ensure the use of distinct task pairs appeared in the long sequence.
  }
  \vspace{7pt}
  \label{tab:long-seq-res}
  \resizebox{0.75\textwidth}{!}{
    \begin{tabular}{clccccc}
      \toprule
       & \multirow{2.5}{*}{\textbf{Method}} & \multicolumn{5}{c}{\textbf{Task Sequence Length}} \\
      \cmidrule(lr){3-7}
      & & \textbf{A$\rightarrow$B} & \textbf{A$\rightarrow$B$\rightarrow$C} &
        \textbf{A$\rightarrow$\dots$\rightarrow$D} & \textbf{A$\rightarrow$\dots$\rightarrow$E} & \textbf{A$\rightarrow$\dots$\rightarrow$F} \\
      \midrule
      \multirow{3}{*}{LS1}
        & $\pi_0$   & 0.0            & 0.0           & 0.0           & 0.0           & 0.0           \\
        & OpenVLA-OFT & 0.0            & 0.0           & 0.0           & 0.0           & 0.0           \\
        & \ours                          & \textbf{100.0} & \textbf{83.3} & \textbf{83.3} & \textbf{75.0} & \textbf{50.0} \\
      \cmidrule(lr){1-7}
      \multirow{3}{*}{LS2}
        & $\pi_0$   & 0.0            & 0.0           & 0.0           & 0.0           & 0.0           \\
        & OpenVLA-OFT & 50.0            & 0.0           & 0.0           & 0.0           & 0.0           \\
        & \ours                          & \textbf{100.0} & \textbf{66.7} & \textbf{66.7} & \textbf{58.3} & 0.0 \\
      \cmidrule(lr){1-7}
      \multirow{3}{*}{LS3}
        & $\pi_0$   & 0.0            & 0.0           & 0.0           & 0.0           & 0.0           \\
        & OpenVLA-OFT & 58.3            & 8.3           & 0.0           & 0.0           & 0.0           \\
        & \ours                          & \textbf{66.7} & \textbf{50.0} & \textbf{41.7} & \textbf{33.3} & 0.0 \\
      \bottomrule
    \end{tabular}%
  }

\end{table}

\subsection{Additional Long Sequence Evaluations}
\label{sec:appendix-exp-long-details}
In Section~\ref{sec:exp}, we evaluate mid-phase task-switching performance in simulation for task LS1 (see Table~\ref{tab:long-seq-task} for task selection). While all baseline methods achieve $0.0\%$ success rates—failing immediately during the initial S2$\rightarrow$S5 transition—\ours demonstrates robust performance. To further validate our approach, we test on two additional long-sequence tasks (LS2 and LS3). As shown in Table~\ref{tab:long-seq-res}, \ours consistently outperforms baselines, achieving substantially higher success rates on these challenging consecutive switching tasks. This further highlights our method's robustness in dynamic task-switching scenarios.


\end{document}